\documentclass[a4paper]{article}

\usepackage{INTERSPEECH2022}
\usepackage{url}

\title{Transsion's Speaker-Attributed Multilingual ASR System for the MLC-SLM 2026 Challenge}
\name{Zhecheng Ren$^1$, Xuanji He$^1$, Xiaoxiao Li$^1$, Zhichen Han$^1$, Gaoyang Dong$^1$, Gaosheng Zhang$^1$, Minchuan Chen$^1$, Fengjie Zhu$^1$}
\address{$^1$Shenzhen Transsion Holdings Co., Ltd, China}
\email{\{zhecheng.ren, xuanji.he, xiaoxiao.li7, zhichen.han, gaoyang.dong, gaosheng.zhang, minchuan.chen, fengjie.zhu\}@transsion.com}

\begin{document}

\maketitle

\begin{abstract}
This paper presents the Transsion Speech Team submission to Task 1 of the MLC-SLM 2026 Challenge, which focuses on speaker-attributed transcription for multilingual conversational speech. We propose a cascaded framework consisting of three components: a speaker diarization module, a long-form multilingual ASR module, and a speaker-transcription fusion module. The diarization module is built upon DiariZen and produces speaker-homogeneous segments through local speaker activity estimation and global speaker clustering. The ASR module is based on Qwen3-Omni and generates multilingual transcriptions, while an external CTC-based alignment model provides precise word- and character-level timestamps. Finally, the fusion module combines diarization outputs with timestamped transcriptions to generate speaker-attributed STM outputs. Experimental results on the official evaluation set demonstrate the effectiveness of the proposed framework. The submitted system achieves a tcpMER of 15.41\% and ranks second among all participating teams.
\end{abstract}

\noindent\textbf{Index Terms}: multilingual ASR, speaker diarization, speaker-attributed transcription, CTC alignment

\section{Introduction}
\label{sec:introduction}

Recent advances in large language models (LLMs) have significantly improved speech-language modeling and multilingual automatic speech recognition (ASR)~\cite{chu2023qwen,chu2024qwen2,tang2023salmonn,zhang2023speechgpt}. By leveraging large-scale pretrained speech encoders and powerful language modeling capabilities, modern multilingual ASR systems achieve strong performance across diverse languages and acoustic conditions~\cite{radford2023robust,peng2024owsm}.

Despite these advances, most existing improvements have been demonstrated under utterance-level settings, where oracle segmentation is available and speaker attribution is not part of the task. Multilingual conversational speech, however, presents additional challenges, including spontaneous speaking styles, rapid turn-taking, short pauses, and speaker alternation under imperfect segmentation~\cite{castillo2025turntaking}. The difficulty is further amplified when speaker diarization and ASR are evaluated jointly, since errors from either component can become coupled in the final speaker-attributed transcription. Consequently, strong utterance-level ASR performance alone does not necessarily translate into strong performance on multilingual conversational diarization and recognition.

Motivated by these challenges, Task 1 of the 2nd Challenge and Workshop on Multilingual Conversational Speech Language Model (MLC-SLM)~\cite{mlcslm2026} focuses on multilingual conversational speech diarization and recognition without oracle segmentation or speaker labels. The official evaluation adopts a time-constrained minimum-permutation mixed error rate (tcpMER) metric, using tcpWER for most languages and tcpCER for Japanese, Korean, and Thai~\cite{mlcslm2026,neumann2023meeteval}. Under this evaluation protocol, a successful system must jointly optimize transcription accuracy and speaker consistency at the conversation level.

In this paper, we present the Transsion Speech Team submission to Task 1 of the MLC-SLM 2026 Challenge. Our approach adopts a cascaded framework consisting of three major components: a speaker diarization module, a long-form multilingual ASR module, and a speaker-transcription fusion module. The diarization module is built upon the DiariZen framework~\cite{han2025diarizen} to produce globally consistent speaker-labeled segments. The long-form ASR module includes audio segmentation, Qwen3-Omni-based transcription~\cite{xu2025qwen3omni}, and CTC-based alignment for fine-grained word-level or character-level timestamps. The resulting diarization outputs and timestamped transcriptions are integrated by a speaker-transcription fusion module to generate the final speaker-attributed STM outputs.

Experimental results on the official MLC-SLM 2026 Task 1 evaluation set demonstrate the effectiveness of the proposed approach. The submitted system achieves a tcpMER of 15.41\% and ranks second among all participating teams.

The remainder of this paper is organized as follows. Section~\ref{sec:datasets} provides an overview of the datasets used in the challenge. Section~\ref{sec:system-description} describes the proposed system in detail, including the speaker diarization, ASR, alignment, and fusion components. Section~\ref{sec:experiments} presents the implementation details and ablation studies. Finally, Section~\ref{sec:conclusion} concludes the paper.

\begin{figure*}[t]
  \centering
  \includegraphics[width=\textwidth,height=0.23\textheight,keepaspectratio]{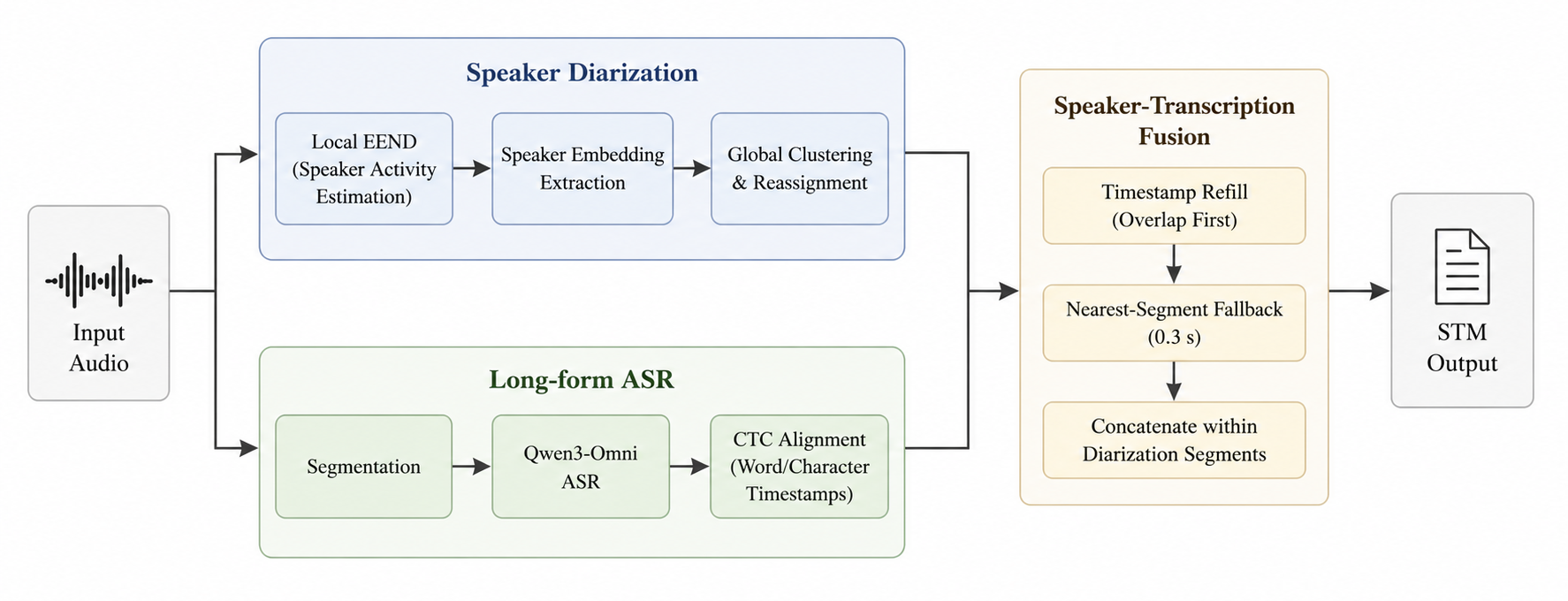}
  \caption{Overview of the proposed cascaded pipeline for multilingual conversational diarization-ASR\@. The system first decomposes the task into two parallel streams: a diarization stream that recovers speaker-consistent conversation structure, and a long-form ASR stream that generates transcriptions together with timing information. The diarization stream proceeds from local speaker activity estimation to speaker embedding extraction and global clustering/reassignment. On the other hand, the ASR stream converts the recording into decoding chunks, transcribes them, and derives word- or character-level timestamps through CTC-based alignment. These two streams are finally reconciled in the fusion stage. There, timestamped ASR outputs are reassigned to the diarization timeline by overlap-first refill with nearest-segment fallback, producing the final STM output.}
  \label{fig:pipeline}
\end{figure*}

\section{Datasets}
\label{sec:datasets}

The official MLC-SLM 2026 Task 1 training set contains approximately 2100 hours of multilingual conversational speech covering 14 languages: English (en), French (fr), German (de), Italian (it), Portuguese (pt), Spanish (es), Japanese (jp), Korean (ko), Russian (ru), Thai (th), Vietnamese (vi), Tagalog (tl), Urdu (ur), and Turkish (tr). At the variant level, the corpus comprises 21 language variants, including five English variants and two variants each for French, Spanish, and Portuguese. Each variant contributes approximately 100 hours of conversational speech. The recordings consist of natural conversations on randomly assigned topics, collected in quiet indoor environments using mobile devices~\cite{mlcslm2026}.

The official dataset provides transcriptions, timestamps, and speaker annotations. Speaker annotations and timestamps are used for training the speaker diarization module, while the transcriptions and timestamps are used to construct the long-form ASR training data described in Section~\ref{sec:training-data-preparation}.

To improve the robustness and generalization of the speaker diarization module, we additionally incorporate several public diarization corpora during training, including AliMeeting~\cite{yu2022m2met}, AISHELL-4~\cite{fu2021aishell4}, NOTSOFAR-1~\cite{vinnikov2024notsofar1}, MagicData-RAMC~\cite{yang2022magicdata}, VoxConverse~\cite{chung2020spot}, and TidyVoice~\cite{farhadipour2026tidyvoice}. These datasets provide diverse recording conditions, speaker characteristics, and conversational scenarios, which help improve diarization performance on multilingual conversational speech.

\section{System Description}
\label{sec:system-description}

Our system consists of three components: speaker diarization (SD), long-form multilingual ASR with CTC-based alignment, and speaker-transcription fusion. Given an input recording, the SD and ASR modules operate in parallel to generate speaker information, transcriptions, and timestamps. Their outputs are subsequently combined in the fusion module to produce the final speaker-attributed STM output. The overall pipeline is illustrated in Figure~\ref{fig:pipeline}.

\subsection{Speaker Diarization}

The speaker diarization module is based on the DiariZen framework~\cite{han2025diarizen,song25c_interspeech}, which follows a local-to-global diarization paradigm. Specifically, local speaker activity is first estimated on short audio chunks, and the resulting local speaker streams are then associated across chunks to derive recording-level speaker identities.

For local diarization, we adopt WavLM-Large~\cite{chen2022wavlm} as the self-supervised pre-trained front-end and use a Conformer encoder~\cite{gulati2020conformer} to produce frame-level speaker activity predictions. To improve the modeling of overlapped speech, the local diarization model is trained with the powerset multi-class cross entropy loss~\cite{plaquet2023powerset}, which treats each possible active-speaker combination as a distinct class.

After local speaker streams are obtained, speaker embeddings are extracted from the corresponding local speaker segments using a speaker embedding extractor~\cite{wang2023campp,lin2024voxblink2}. The extracted embeddings are then clustered with VBx~\cite{palka2026vbx} to merge local speaker tracks into globally consistent speaker identities. Finally, the clustering results are used to reassign local speaker streams and generate the final diarization outputs for downstream transcription fusion.

\subsection{Long-form ASR}

The long-form multilingual ASR module operates on a full conversation recording and produces both its transcription and the associated timestamps. It consists of three submodules: segmentation, ASR, and CTC-based alignment. The segmentation submodule converts the full recording into shorter audio chunks, while preserving sufficient contextual information for robust transcription. The ASR submodule then transcribes these chunks with a Qwen3-Omni-based model~\cite{xu2025qwen3omni} to generate chunk-level transcription hypotheses. The CTC-based alignment submodule is built on Qwen3-ASR-1.7B~\cite{ma2025qwen3asr} and serves as the timestamping model in the pipeline. It processes each audio chunk together with its predicted text and produces word- or character-level timestamps. The resulting transcriptions and timestamps are then forwarded to the downstream speaker-transcription fusion module.

\subsection{Speaker-Transcription Fusion}
\label{sec:fusion}

The speaker-transcription fusion module integrates the outputs of the two upstream modules to generate speaker-attributed transcriptions. Specifically, the diarization module provides speaker-labeled segments, while the CTC-based alignment module generates word-level or character-level timestamps for the ASR output. During the refill stage, the timestamped words or characters are mapped back to the diarization timeline. Each word or character is first assigned to a diarization segment based on temporal overlap. If no overlapping segment is found, a nearest-segment fallback strategy is applied, where the token is assigned to the closest diarization segment within a 0.3-second tolerance window. After the reassignment process, all words or characters belonging to the same diarization segment are concatenated according to their temporal order to form the corresponding transcription. The resulting speaker-attributed transcriptions are then converted into the STM format for evaluation.

\section{Experiments}
\label{sec:experiments}

This section presents the system implementation details and the ablation studies used to analyze the main design choices of our approach. We first describe the data preparation pipeline, training configurations, and inference-time implementation details. The discussion focuses on the final submitted system, while also covering alternative configurations used in the ablation experiments. We then evaluate the contribution of several major components through controlled ablation experiments.

\subsection{System Implementation Details}

\subsubsection{Training Data Preparation}
\label{sec:training-data-preparation}

The official training recordings contain a considerable amount of speech that is not covered by the provided annotations. To avoid introducing supervision noise, these unlabeled speech regions are first muted during data preparation. The resulting recordings are used for training the speaker diarization module.

For the Qwen3-Omni ASR module, additional processing is applied to construct training samples. Specifically, each recording is first segmented according to the speaker annotations, and neighboring segments separated by less than 0.5\,s are concatenated. This procedure preserves conversational context while maintaining a manageable input length for model training.

\subsubsection{Speaker Diarization}

The local EEND model is trained for 30 epochs. WavLM is frozen during the first two epochs and then jointly optimized with the diarization backend. The learning rates are set to $1\times10^{-5}$ for WavLM and $1\times10^{-3}$ for the remaining components. To improve robustness, noise augmentation is progressively introduced, with single-noise augmentation starting from epoch 6 and mixed-noise augmentation from epoch 15.

For global speaker labeling, we evaluate different speaker embedding extractors and PLDA backends in the speaker clustering stage. Experimental results show that VoxBlink2 embeddings~\cite{lin2024voxblink2} consistently outperform CAM++ embeddings~\cite{wang2023campp} when combined with the same PLDA backend. Furthermore, the multilingual PLDA model trained on the TidyVoice corpus provides additional improvements over the original PLDA backend~\cite{kenny2010plda}. Among all evaluated configurations, the combination of VoxBlink2 embeddings and the TidyVoice-trained PLDA backend achieves the strongest diarization performance and is therefore adopted in the final submitted system.

\subsubsection{Long-form ASR}

\textbf{Segmentation:}
We evaluate two segmentation strategies during inference: SD-conditioned segmentation and VAD-based segmentation. The final submitted system adopts the former.

To better preserve conversational context and align the inference distribution with the ASR training data, the SD-conditioned strategy first partitions each recording according to the speaker boundaries predicted by the diarization module. Adjacent segments are then greedily merged into longer chunks. The merging process is constrained by a maximum inter-segment gap of 0.5\,s and a target chunk duration range of 1--30\,s.

As an alternative, we also investigate a VAD-based strategy using a FunASR FSMN VAD model~\cite{gao2023funasr}. Consecutive speech regions separated by less than 0.1\,s are merged, while segments longer than 30\,s are split. To reduce excessive fragmentation, short segments are further merged with neighboring segments when the inter-segment gap is below 0.5\,s.

\textbf{Qwen3-Omni ASR:}
The ASR backend is built upon Qwen3-Omni-30B-A3B-Instruct~\cite{xu2025qwen3omni} and adapted using Low-Rank Adaptation (LoRA)~\cite{hu2022lora}. Training is conducted in two stages.

In the first stage, the model is trained on the prepared training data described in Section~\ref{sec:training-data-preparation} using LoRA rank 8, $\alpha=32$, and zero dropout. The LoRA adapters are inserted into both the attention and feed-forward projections, while the vision encoder and aligner remain frozen throughout training. The checkpoint with the best validation performance is selected to initialize the second stage.

The second stage aims to improve underperforming language variants through continued training on a mixture of the original training data and selectively speed-perturbed samples. The LoRA rank is increased to 32 to provide additional adaptation capacity. The final ASR model is obtained by averaging multiple complementary checkpoints from the second stage and merging the resulting adapters into the first-stage model.

During inference, the final ASR model generates chunk-level transcriptions using the language-conditioned prompt \textit{``Transcribe the speech in \{language\}. Output only the transcript.''} with greedy decoding. The resulting hypotheses are subsequently passed to the CTC alignment and speaker-transcription fusion modules.

\textbf{CTC Alignment:}
The alignment module is built upon Qwen3-ASR-1.7B~\cite{ma2025qwen3asr} and trained separately from the Qwen3-Omni ASR model. An auxiliary CTC head is attached to the audio encoder and jointly optimized with the autoregressive objective using loss weights of 0.3 and 1.0, respectively. CTC targets are derived from the Whisper tokenizer~\cite{radford2023robust} with an additional blank symbol.

At inference time, the model produces word-level timestamps for space-delimited languages and character-level timestamps for Japanese, Korean, and Thai. Failed alignments are handled by timestamp interpolation, while hallucinated repetitions are removed and realigned. The resulting timestamps are used by the downstream speaker-transcription fusion module.

\subsubsection{Speaker-Transcription Fusion}

The final submitted system adopts the overlap-first fusion strategy described in Section~\ref{sec:fusion}, with nearest-segment fallback for unmatched tokens. The reassigned words or characters are concatenated within each diarization segment to generate the final STM output.

For comparison, we also evaluate a midpoint-based strategy that assigns each token according to the midpoint of its CTC-derived timestamp interval.

\subsection{Ablation Studies}

We conduct incremental ablation experiments to evaluate the contribution of each major component in the proposed system. Starting from a baseline configuration, we progressively introduce (1) task-specific Local EEND fine-tuning, (2) the overlap-first speaker-transcription fusion with nearest-segment fallback strategy, (3) VoxBlink2 embeddings with a TidyVoice-trained PLDA backend, and (4) SD-conditioned segmentation. The results are summarized in Table~\ref{tab:ablation}.

Our baseline system is constructed using the open-source DiariZen diarization model without task-specific fine-tuning, the original CAM++ speaker embedding extractor and PLDA backend, a VAD-based segmentation strategy, and a midpoint-based speaker-transcription fusion method. The Qwen3-Omni ASR model and the CTC alignment model remain unchanged throughout all ablation experiments.

As an additional reference, we also fine-tune VibeVoice-ASR~\cite{peng2026vibevoice}, a fully end-to-end model that jointly performs speaker diarization and speech recognition. The resulting system achieves a tcpMER of 16.27\% on the official evaluation set, providing a representative end-to-end baseline for comparison with our cascaded approach.

\begin{table}[t]
\centering
\small
\caption{Reference result and incremental ablation results on the official evaluation set.}
\label{tab:ablation}
\begin{tabular}{lc}
\toprule
Configuration & tcpMER (\%) \\
\midrule
Fine-tuned VibeVoice-ASR & 16.27 \\
\midrule
Our baseline & 22.78 \\
+ Fine-tuned local EEND & 16.40 \\
+ Overlap-first fusion with fallback & 16.01 \\
+ VoxBlink2 + TidyVoice-trained PLDA & 15.72 \\
+ SD-conditioned segmentation & \textbf{15.41} \\
\bottomrule
\end{tabular}
\end{table}

\subsubsection{Local EEND}

We first evaluate the impact of task-specific Local EEND fine-tuning. Compared with the open-source DiariZen model, the fine-tuned EEND substantially improves speaker activity estimation and reduces tcpMER from 22.78\% to 16.40\%.

\subsubsection{Speaker-Transcription Fusion}

We then replace the midpoint-based fusion strategy with the proposed overlap-first strategy with nearest-segment fallback. This modification improves token assignment robustness and further reduces tcpMER from 16.40\% to 16.01\%.

\subsubsection{Speaker Embedding Extractor and PLDA Backend}

Replacing CAM++ and the original PLDA backend with VoxBlink2 and multilingual TidyVoice-trained PLDA improves global speaker clustering and reduces tcpMER from 16.01\% to 15.72\%.

\subsubsection{Segmentation}

Finally, SD-conditioned segmentation is adopted in place of VAD-based segmentation. By producing ASR chunks that better match the training distribution, this strategy further reduces tcpMER from 15.72\% to 15.41\%.

\section{Conclusion}
\label{sec:conclusion}

This paper presents the Transsion Speech Team submission to Task 1 of the MLC-SLM 2026 Challenge. We propose a cascaded framework that combines speaker diarization, long-form multilingual ASR, CTC-based alignment, and speaker-transcription fusion to generate speaker-attributed transcriptions for multilingual conversational speech.

Ablation studies demonstrate the effectiveness of task-specific EEND fine-tuning, multilingual speaker clustering, SD-conditioned segmentation, and overlap-first fusion. On the official evaluation set, the proposed system achieves a tcpMER of 15.41\% and ranks second among all participating teams.

\bibliographystyle{IEEEtran}
\bibliography{references}

\end{document}